\pdfoutput=1

\documentclass[11pt]{article}

\usepackage{acl}

\usepackage{times}
\usepackage{latexsym}

\usepackage[T1]{fontenc}

\usepackage[utf8]{inputenc}

\usepackage{microtype}
\usepackage{graphicx}
\usepackage{subcaption}
\usepackage{booktabs}
\usepackage{multirow}
\usepackage{amsmath}

\newif\ifshowcomments
\showcommentstrue
\ifshowcomments
\newcommand{\mynote}[2]{\textcolor{blue}{\fbox{\bfseries\sffamily\scriptsize#1}}
  \textcolor{blue}{{$/*$\textsf{\emph{#2}}$*/$}}}
\ifshowcomments

\ifshowcomments

\ifshowcomments

\else
\newcommand{\mynote}[2]{}
\fi

\usepackage{textcomp}
\usepackage{xcolor}

\title{Online Personalizing White-box LLMs Generation with
Neural Bandits}

\author{\hspace{-3cm}Zekai Chen \\ \hspace{-3cm}J.P. Morgan Chase \\ \hspace{-3cm}New York, NY \\ \hspace{-3cm}{\small \texttt{zekai.chen@jpmchase.com}}
        \And
        \hspace{-2cm}Weeden Daniel~~~Po-yu Chen~~~Francois Buet-Golfouse \\
\hspace{-2cm}J.P. Morgan Chase \\ \hspace{-2cm}London, UK \\ \hspace{-2cm}{\small                  \texttt{francois.buet-golfouse@jpmorgan.com}}
        }

\begin{document}
\maketitle

\begin{abstract}
The advent of personalized content generation by LLMs presents a novel challenge: how to efficiently adapt
text to meet individual preferences without the unsustainable demand of creating a unique model for each user. This study introduces an innovative online method that employs neural bandit algorithms to dynamically optimize soft instruction embeddings based on user feedback, enhancing the personalization of open-ended text generation by white-box LLMs. Through rigorous experimentation on various tasks, we demonstrate significant performance improvements over baseline strategies. NeuralTS, in particular, leads to substantial enhancements in personalized news headline generation, achieving up to a 62.9\% improvement in terms of best ROUGE scores and up to 2.76\% increase in LLM-agent evaluation against the baseline. 
\end{abstract}

\section{Introduction}
\label{sec:intro}

In recent years, the advancements in large language models (LLMs) have been remarkable~\citep{Brown2020LanguageMA,Zhao2023ASO}, with these models demonstrating an unparalleled ability to understand and generate text across a wide spectrum of tasks~\citep{Wei2022ChainOT,Kojima2022LargeLM}. This capability has 
revolutionized the way we interact with machine-generated content 
and 
opened up new avenues for personalized text generation~\citep{Kirk2023PersonalisationWB,Li2023TeachLT}. 

Personalization in text generation is of paramount importance 
to ensure
user engagement and satisfaction~\citep{Huang2022UserNLP222I}
across a range of in applications such as composing tweets, or generating news articles and financial reports, or in more personalized settings like business communications and creative writing~\citep{Li2019TowardsCA,Li2020KnowledgeEnhancedPR}.
\begin{figure}[]
    \centering
    \includegraphics[width=\linewidth]{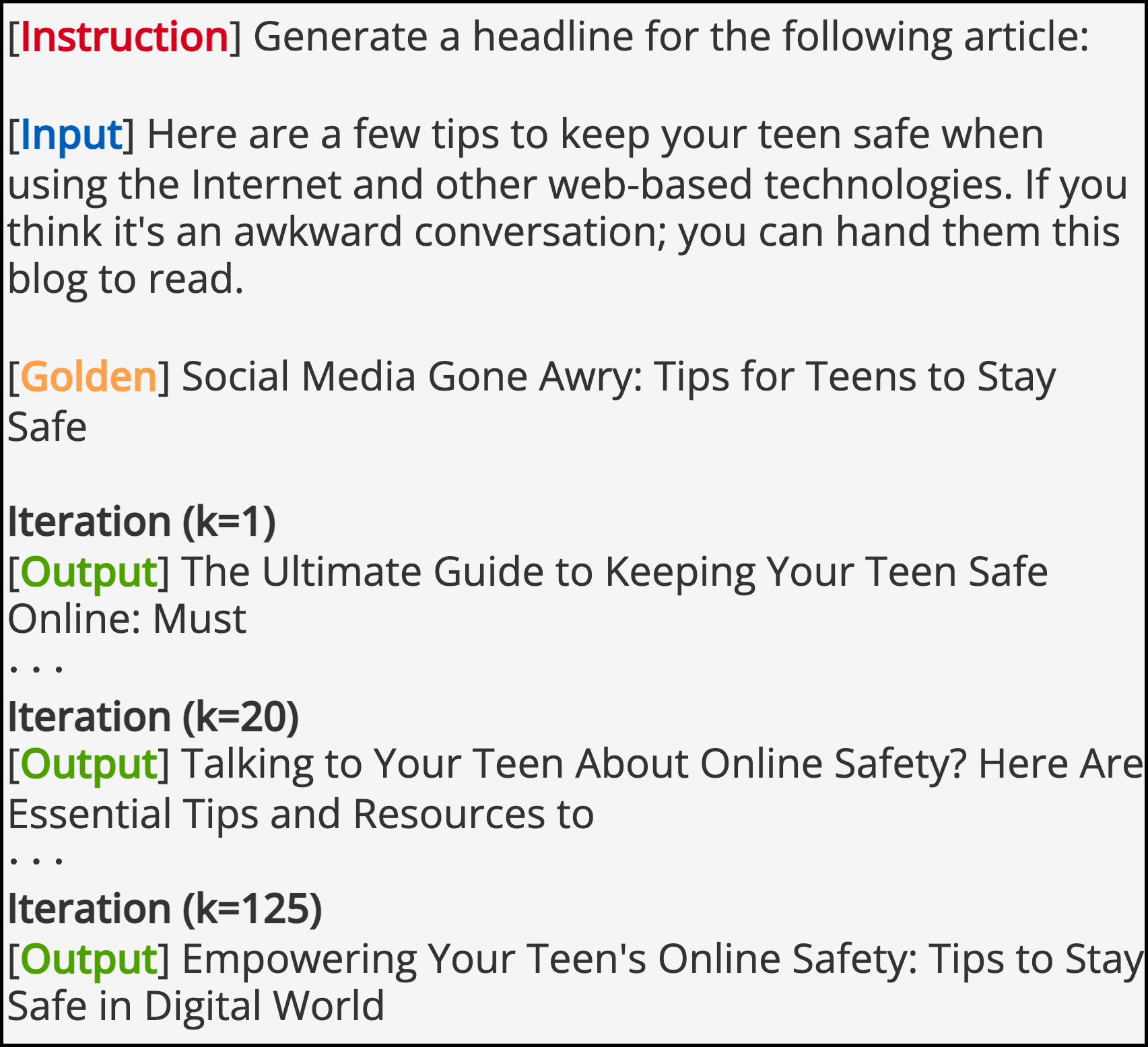}
    \caption{Evolution of generated headlines for an article on teen internet safety, illustrating the progressive refinement of generation that emulates this journalist stylistic tendencies through online learning.}
    \vspace{-3mm}
    \label{fig:lamp_example}
\end{figure}

\begin{figure*}[h!]
    \centering
    \vspace{-5mm}
    \includegraphics[width=.65\linewidth]{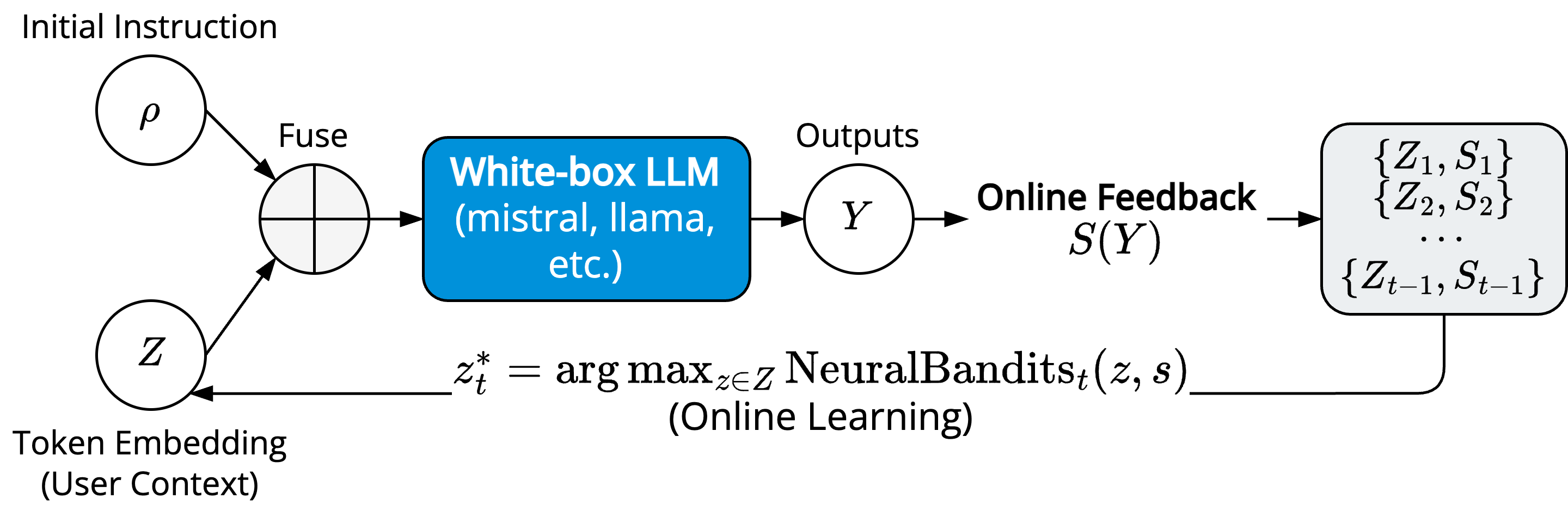}
    \caption{Illustration of our framework. Details are described in Section~\ref{sec:method}.}
    \label{fig:main}
\end{figure*}

However, the prospect of developing a unique LLM for each user presents 
challenges, including the prohibitive resource requirements~\citep{Hoffmann2022TrainingCL}, data privacy concerns~\citep{Li2023MultistepJP}, and the scarcity of personalized data~\citep{Rafailov2023DirectPO}. These obstacles necessitate an alternative strategy that is both practical and flexible. A promising solution lies in adopting lightweight models capable of online learning, which can dynamically adjust their output based on continuous user feedback~\citep{Bai2022TrainingAH}. Such an approach not only circumvents the need for a bespoke model for each user but also encourages 
alignment of the 
generated content 
to individual preferences over time. Importantly, this adaptive process is poised to unlock long-term rewards stemming from personalization, encompassing not just explicit preferences expressed by users but also responding to favorable actions~\citep{Xie2021InteractionGroundedL}. 

Despite these benefits, the ultimate effectiveness of LLMs hinges on the quality of the given instructions~\citep{Zhou2022LargeLM,Bang2023AMM,White2023APP}. Previous efforts have focused on gradient-based strategies~\citep{Shin2020ElicitingKF,Li2021PrefixTuningOC,Lester2021ThePO} for automated instruction optimization, the applicability is limited to less advanced public models, leaving out many advanced yet proprietary models. With the emergence of more advanced open models such as Mistral-7B~\citep{Jiang2023Mistral7}, Llama-70B~\citep{Touvron2023LLaMAOA,Touvron2023Llama2O}, and Mixtral-8x7B~\citep{Jiang2024MixtralOE}, which offer transparency and have reported performance that even surpasses that of ChatGPT-3.5\footnote{https://chat.openai.com/}, there is a renewed focus on leveraging these models for direct optimization. 


In this study, we introduce a novel online method for enhancing the personalization of open-ended text generation with white-box LLMs. Considering that {\itshape capturing the nuances of persona in natural language instructions is challenging}, we aim to directly optimize the soft token embeddings~\citep{Chen2023InstructZeroEI,Lin2023UseYI}, representing the contextual factors through user feedback by utilizing neural bandit algorithms~\citep{Zhou2019NeuralCB,zhang2021neural}. This method not only promises to refine the personalization process of text generation but also contributes to the broader application of adaptive algorithms in creating content that closely reflects individual user preferences.

\section{Peronalization with Neural Bandits}
\label{sec:method}

\begin{figure*}
    \centering
    \vspace{-5mm}
    \includegraphics[width=\linewidth]{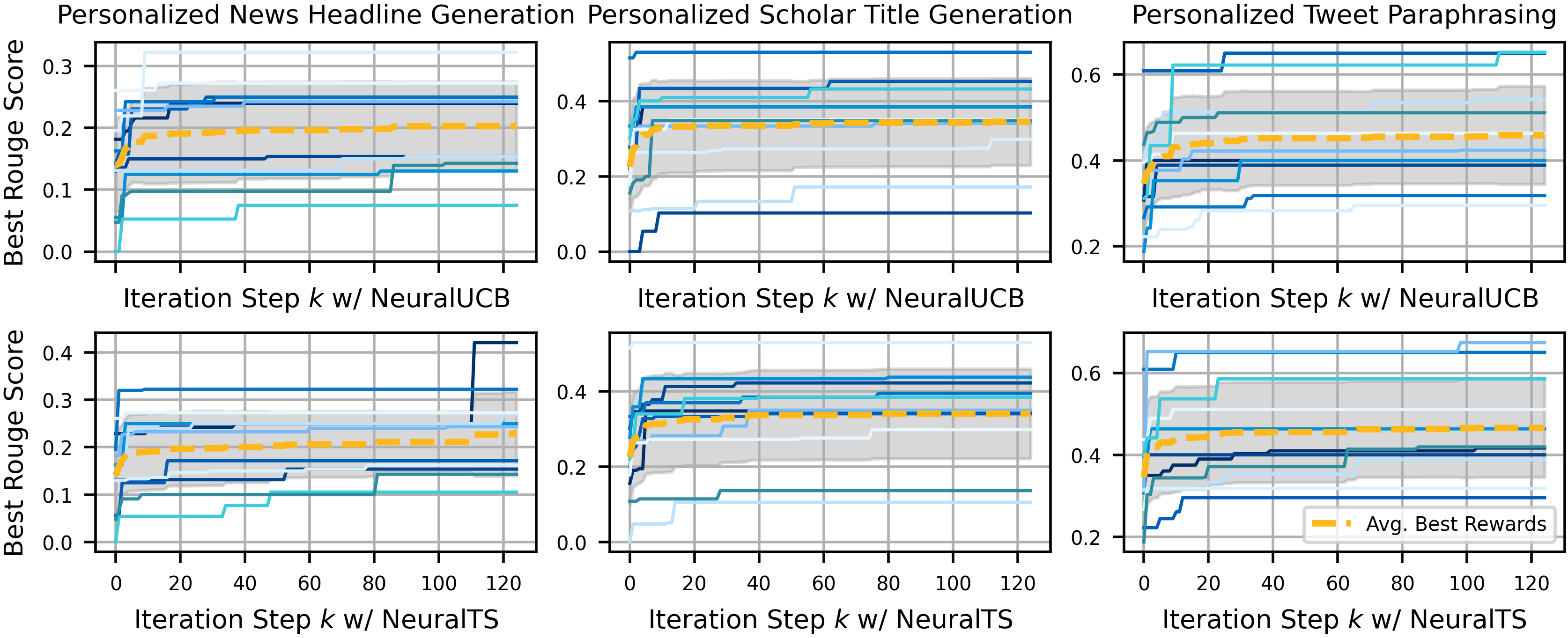}
    \caption{10 user profiles (different blues) are randomly selected for demonstration. Trend of increasing averaged best rewards (yellow dashes) across learning iterations for three personalized text generation tasks, showcasing the progressive improvement in performance achieved by both NeuralUCB~\citep{Zhou2019NeuralCB} and NeuralTS~\citep{zhang2021neural} algorithms.}
    \label{fig:res}
\end{figure*}

Neural Bandits~\citep{Zhou2019NeuralCB,zhang2021neural} integrate the adaptive exploration of bandit algorithms~\citep{Auer2003UsingCB,Agrawal2012ThompsonSF,Li2010ACA} with neural networks' superior ability to predict rewards under uncertainty. By leveraging past interactions to balance the trade-off between exploring new actions and exploiting known ones, these algorithms can accurately predict and enhance personalized outcomes. Therefore, we adopt NeuralUCB~\citep{zhou2020neural} and NeuralTS~\citep{zhang2021neural} in our framework to comprehensively evaluate how Neural Bandits benefit LLMs' generation in an online fashion. 


This section explores the application of Neural Bandits to white-box large language models (LLMs) as a strategy for directly refining soft prompts ({\itshape aka.} contextual embeddings) to overcome the \emph{inability of natural language instructions to fully express nuances of persona}.
The process involves a white-box LLM, \(f\), which takes a soft prompt \(z\), an instruction, \(\rho\), and a test input, \(x\), to produce an output sentence, \(\hat{y} = f(z, \rho, x)\). A soft prompt \(z\) is a continuous vector representing preference token embeddings, which is used alongside the default instruction, \(\rho\), as input to \(f\). Given a specific user \(u\) at time \(t\), we have a specific input \(x_i\), the goal is to find an optimal soft prompt token, \(z^*\), that maximizes the following objective function based on user's feedback \(s(\hat{y}_i)\) (see Figure \ref{fig:main} for illustration):
\[z_{t}^* = \arg \max_{z \in Z} \textrm{NeuralBandits}_{t}(z, s)\]
With NeuralUCB, it involves computing the acquisition value for each candidate soft prompt and selecting the one that maximizes this value, based on the model's current parameters and the uncertainty associated with each prompt:
\[z_{t+1} = \arg\max_{z \in Z} \left(\mu(g(z); \theta_{t}) + \nu_t \sigma_{t}(g(z); \theta_{t}) \right),
\]
Here, \(\mu(g(z); \theta_{t})\) represents the network's predicted value for the soft prompt \(z\) at iteration \(t\), while \(\sigma_{t}(g(z); \theta_{t})\) quantifies the uncertainty of the prediction. The parameter \(\nu_t\) controls the trade-off between exploration and exploitation, influencing the algorithm's preference for exploring less certain prompts versus exploiting prompts with higher predicted values.

When using NeuralTS for updating \(z\), the process involves sampling from a predictive distribution to select the next soft prompt. Unlike NeuralUCB, which directly uses a deterministic acquisition function, NeuralTS generates a sample for each candidate \(z\) from a distribution modeled by the neural network. The update can be represented as follows:

\[z_{t+1} = \arg\max_{z \in Z} \tilde{r}_{z,t}\]
\[\tilde{r}_{z,t}  \sim \mathcal{N}\Big(\tilde{\mu}\big(g(z); \theta_{t}\big), \nu_t \tilde{\sigma}_{t}\big(g(z); \theta_{t}\big)\Big)\]

Where, \(\tilde{\mu}(g(z); \theta_{t})\) and \(\tilde{\sigma}_{t}(g(z); \theta_{t})\) 
represent the estimated mean and standard deviation respectively for the soft prompt \(z\), and \(\tilde{r}_{z,t}\) is a sampled reward from the predictive distribution for \(z\) at iteration \(t\). The parameter \(\nu_t\) again balances exploration and exploitation, but in the context of NeuralTS, the exploration is informed by the stochasticity introduced through sampling, encouraging diversity in the selection of \(z\) based on both prediction and uncertainty.


NeuralUCB uses an upper confidence bound to balance these aspects deterministically, offering robust performance in environments where a clear quantification of uncertainty benefits decision-making. NeuralTS, on the other hand, employs a Thompson sampling~\citep{Agrawal2012ThompsonSF} approach, introducing stochasticity in the selection process, which can lead to more diverse exploration. Therefore, we tested both approaches as different kernels given the actual performance hinges on the problem's nature, the desired balance between exploration and exploitation, and the computational resources available.

\section{Experiments on LaMP}

\subsection{Personalized Generations}
The LaMP~\cite{salemi2023lamp} dataset, or Language Model Personalization, is a benchmark designed for training and evaluating large language models (LLMs) to produce personalized outputs. It aims to assess the efficacy of LLMs in generating responses tailored to individual user profiles. For example, in personalized news headline generation task, the writing of journalist exhibits unique stylistic characteristics shaped by both individual and societal influences~\cite{Zhu2021IdiosyncraticBN}. This scenario serves as an excellent opportunity for exploring personalized generation.


In this study, we evaluate our framework on three open-ended generation tasks which are Personalized News Headline Generation, Personalized Scholarly Title Generation, and Personalized Tweet Paraphrasing. 

\begin{table*}[h!]
\centering
\vspace{-3mm}
\resizebox{0.75\linewidth}{!}{%
\begin{tabular}{@{}llcccl@{}}
\toprule
Tasks & Metric & \begin{tabular}[c]{@{}c@{}}Random\\ \(\textrm{(zero-shot)}\)\end{tabular} & \begin{tabular}[c]{@{}c@{}}NeuralUCB\\ \(@k=165\)\end{tabular} & \begin{tabular}[c]{@{}c@{}}NeuralTS \\ \(@k=165\)\end{tabular} & \(\Delta\uparrow\) \\ \midrule
\begin{tabular}[c]{@{}l@{}}Personalized\\ News Headline Generation\end{tabular} & Avg. ROUGE-1/L\(\uparrow\) & \(0.140\pm 0.076\) & \(0.203\pm 0.072\) & \(\textbf{0.228}\pm \textbf{0.088}\) & \(62.9\%\) \\ \midrule
\begin{tabular}[c]{@{}l@{}}Personalized\\ Scholarly Title Generation\end{tabular} & Avg. ROUGE-1/L\(\uparrow\) & \(0.225\pm 0.127\) & \(\textbf{0.345}\pm \textbf{0.116}\) & \(0.341\pm 0.119\) & \(53.3\%\) \\ \midrule
\begin{tabular}[c]{@{}l@{}}Personalized\\ Tweet Paraphrasing\end{tabular} & Avg. ROUGE-1/L\(\uparrow\) & \(0.346\pm 0.110\) & \(0.459\pm 0.114\) & \(\textbf{0.466}\pm \textbf{0.120}\) & \(34.7\%\) \\ \bottomrule
\end{tabular}%
}
\caption{Performance comparison of NeuralUCB~\citep{Zhou2019NeuralCB} and NeuralTS~\citep{zhang2021neural} algorithms against a random baseline in three personalized text generation tasks, measured by average ROUGE-1/L scores with improvement percentages (\(\Delta\uparrow\)) after 165 iterations.}
\label{tab:main}
\end{table*}

\begin{figure}
    \centering
    \includegraphics[width=\linewidth]{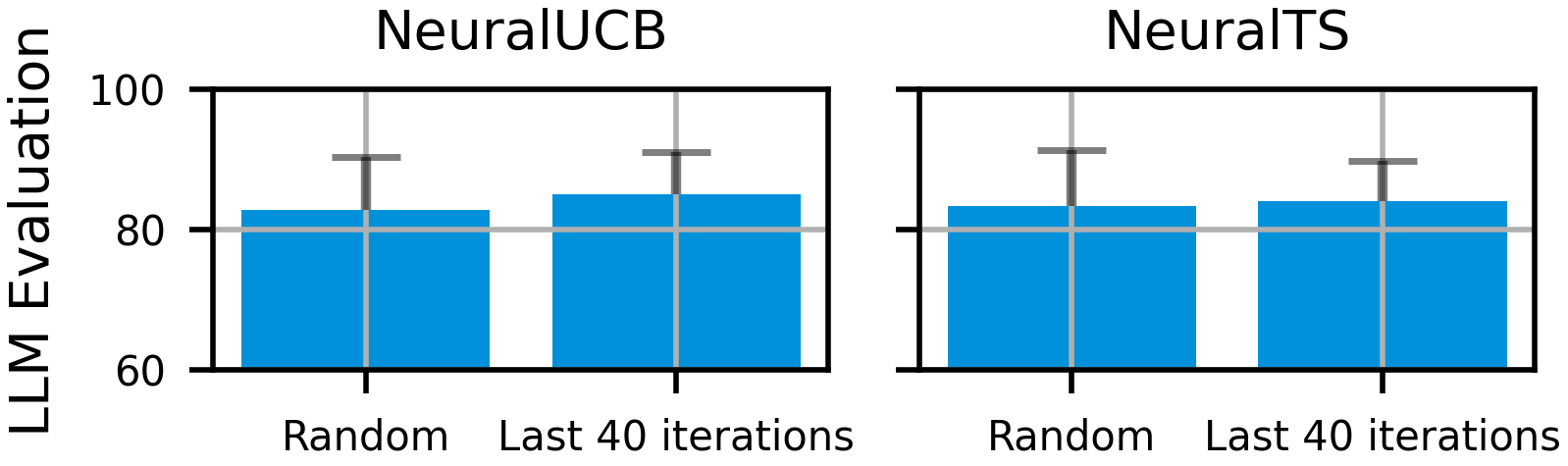}
    \caption{LLM evaluation of personalized generation between NeuralUCB~\citep{Zhou2019NeuralCB} and NeuralTS~\citep{zhang2021neural} in personalized news headline generation.}
    \label{fig:llm-eval}
\end{figure}

\subsection{Online Simulation using LaMP}
We simulate a realistic online setting where individual users, such as those represented by distinct profiles in the LaMP dataset, continually receive outputs from LLMs in response to predetermined instructions. We then take ROUGE scores~\citep{lin2004rouge}, which measure the correspondence between the model's output and the user's ideal or "golden" response, as their {\itshape online feedback}. Additionally, we monitor the assessments from a black-box LLM, which evaluates the appropriateness of the generated content for a desired persona based on their profiles ({\it e.g.}, using LLMs to automatically rate the generation based on the consistency with users' stylistics). An ideal result would exhibit both improved ROUGE metrics and LLM ratings.


As introduced in Section~\ref{sec:intro}, we employ Mistral-7B-Instruct-v0.2\footnote{https://huggingface.co/mistralai/Mistral-7B-Instruct-v0.2} as our default white-box LLM. By simulating this online interaction scenario, our aim is to closely mimic the dynamic and personalized experience users encounter in real-world applications of language models. Through this experiment, we seek to evaluate the effectiveness of Neural Bandits in responding to the nuanced persona of different users/profiles, thereby informing future improvements and adaptations in personalized generation.

\subsection{Results Analysis}

For the task of Personalized News Headline Generation, from Table~\ref{tab:main} we observed a significant performance leap with NeuralTS, achieving an Average ROUGE-1/L score of \(0.228 \pm 0.088\), marking a \(62.9\%\) improvement over the random baseline (\(0.140 \pm 0.076\)). NeuralUCB also outperformed the baseline, albeit with a slightly lower score of \(0.203 \pm 0.072\). In Personalized Scholarly Title Generation, NeuralUCB exhibited the highest increase, with a score of \(0.345 \pm 0.116\), closely followed by NeuralTS (\(0.341 \pm 0.119\)), both surpassing the baseline (\(0.225 \pm 0.127\)) by approximately \(53.3\%\). Lastly, the Personalized Tweet Paraphrasing task showed NeuralTS slightly outperforming NeuralUCB with a score of \(0.466 \pm 0.120\) against \(0.459 \pm 0.114\), over a baseline of \(0.346 \pm 0.110\), translating to a \(34.7\%\) improvement. Figure~\ref{fig:main} also summarises these results. At the same time, as shown in Figure~\ref{fig:llm-eval}, we observe increase over the LLM evaluation scores, with NeuralUCB achieving an average improvement of \(2.8\%\) vs. \(\sim1\%\) for NeuralTS. These findings underscore the potential of Neural Bandit algorithms in enhancing content personalization across various text generation tasks. 


However, it's important to acknowledge certain limitations inherent in LLM evaluations. LLMs tend to exhibit a positivity bias, often returning higher scores for the generated content. This bias can potentially skew the evaluation results and may not fully capture nuances in personalization or consistency. Also, while ROUGE provides a useful automated proxy for assessing the quality of personalized content, it has well-known limitations in fully capturing subjective preferences or nuanced stylistic tendencies. It may not align well with human judgments, especially for highly creative and personalized text. Thus, while the gains demonstrated in both ROUGE and LLM evaluation provide encouraging evidence, human evaluations are imperative for comprehensively assessing improvements in adapting to individual user profiles. 


\section{Conclusion}
In conclusion, this study presents a novel framework for personalized content generation using large language models (LLMs) through the application of neural bandit algorithms. By dynamically optimizing soft instruction embeddings through online feedback, our method demonstrates improvements in personalizing open-ended text generation tasks as scored by both ROUGE scores and an LLM evaluator. Further research is needed to confirm that these results are consistent with human-evaluated suitability.


\section{Limitations}
We discuss some limitations this work may present in this section. The evaluation in this study is limited to only three generation tasks from the LaMP benchmark. Testing on a more diverse and extensive set of personalized taks would strengthen the claims around the general applicability of the method. The simulation of an online environment with user profiles from the LaMP dataset may not fully capture the nuances of real-world user interactions and feedback. Evaluating with actual human subjects over longer time horizons could reveal additional challenges. Additionally, ROUGE score, is used to assess generation quality. Incorporating other metrics and human evaluations could provide a more holistic view of the improvements achieved. As of now, only two neural bandit algorithms, NeuralUCB and NeuralTS, are investigated. Comparing to a wider range of adaptive optimization algorithms could provide deeper insights into the most suitable techniques.

\section{Ethics Consideration}
In real-world, collecting personalized data like past personal data raises privacy concerns around data usage and consent. In our study, we use LaMP which is a public dataset with user consent. To some extent, tailoring content to align with a user's preferences risks creating isolated filter bubbles that entrench viewpoints. This is also an important reason that we investigate bandit algorithms to better balance exploration against exploitation and serve as mitigation strategies which exposes users to diverse perspectives. One potential risk is that the capability to generate highly persuasive personalized text could be misused for deceptive purposes. Safeguards against generating harmful, unethical or untruthful content need to be in place. Like any powerful technology, personalized generation abilities could be co-opted for nefarious ends by malicious actors. Responsible disclosure and governance practices are critical. In summary, the ethical dimensions span privacy, security, fairness, transparency, accountability and the broader social impacts of deploying personalized generative models. A principled, human-centric approach that places ethics at the foundation of research and development is imperative as these capabilities continue to advance.



\bibliography{acl_natbib}

\begin{thebibliography}{36}
\expandafter\ifx\csname natexlab\endcsname\relax\def\natexlab#1{#1}\fi

\bibitem[{Agrawal and Goyal(2012)}]{Agrawal2012ThompsonSF}
Shipra Agrawal and Navin Goyal. 2012.
\newblock \href {https://api.semanticscholar.org/CorpusID:96146} {Thompson sampling for contextual bandits with linear payoffs}.
\newblock In \emph{ICML}.

\bibitem[{Auer(2003)}]{Auer2003UsingCB}
Peter Auer. 2003.
\newblock \href {https://api.semanticscholar.org/CorpusID:10485293} {Using confidence bounds for exploitation-exploration trade-offs}.
\newblock \emph{JMLR}, 3:397--422.

\bibitem[{Bai et~al.(2022)Bai, Jones, Ndousse, Askell, Chen, DasSarma, Drain, Fort, Ganguli, Henighan, Joseph, Kadavath, Kernion, Conerly, El-Showk, Elhage, Hatfield-Dodds, Hernandez, Hume, Johnston, Kravec, Lovitt, Nanda, Olsson, Amodei, Brown, Clark, McCandlish, Olah, Mann, and Kaplan}]{Bai2022TrainingAH}
Yuntao Bai, Andy Jones, Kamal Ndousse, Amanda Askell, Anna Chen, Nova DasSarma, Dawn Drain, Stanislav Fort, Deep Ganguli, T.~J. Henighan, Nicholas Joseph, Saurav Kadavath, John Kernion, Tom Conerly, Sheer El-Showk, Nelson Elhage, Zac Hatfield-Dodds, Danny Hernandez, Tristan Hume, Scott Johnston, Shauna Kravec, Liane Lovitt, Neel Nanda, Catherine Olsson, Dario Amodei, Tom~B. Brown, Jack Clark, Sam McCandlish, Christopher Olah, Benjamin Mann, and Jared Kaplan. 2022.
\newblock \href {https://api.semanticscholar.org/CorpusID:248118878} {Training a helpful and harmless assistant with reinforcement learning from human feedback}.
\newblock \emph{ArXiv}, abs/2204.05862.

\bibitem[{Bang et~al.(2023)Bang, Cahyawijaya, Lee, Dai, Su, Wilie, Lovenia, Ji, Yu, Chung, Do, Xu, and Fung}]{Bang2023AMM}
Yejin Bang, Samuel Cahyawijaya, Nayeon Lee, Wenliang Dai, Dan Su, Bryan Wilie, Holy Lovenia, Ziwei Ji, Tiezheng Yu, Willy Chung, Quyet~V. Do, Yan Xu, and Pascale Fung. 2023.
\newblock \href {https://api.semanticscholar.org/CorpusID:256662612} {A multitask, multilingual, multimodal evaluation of chatgpt on reasoning, hallucination, and interactivity}.
\newblock \emph{IJNLP}, abs/2302.04023.

\bibitem[{Brown et~al.(2020)Brown, Mann, Ryder, Subbiah, Kaplan, Dhariwal, Neelakantan, Shyam, Sastry, Askell, Agarwal, Herbert-Voss, Krueger, Henighan, Child, Ramesh, Ziegler, Wu, Winter, Hesse, Chen, Sigler, Litwin, Gray, Chess, Clark, Berner, McCandlish, Radford, Sutskever, and Amodei}]{Brown2020LanguageMA}
Tom~B. Brown, Benjamin Mann, Nick Ryder, Melanie Subbiah, Jared Kaplan, Prafulla Dhariwal, Arvind Neelakantan, Pranav Shyam, Girish Sastry, Amanda Askell, Sandhini Agarwal, Ariel Herbert-Voss, Gretchen Krueger, T.~J. Henighan, Rewon Child, Aditya Ramesh, Daniel~M. Ziegler, Jeff Wu, Clemens Winter, Christopher Hesse, Mark Chen, Eric Sigler, Mateusz Litwin, Scott Gray, Benjamin Chess, Jack Clark, Christopher Berner, Sam McCandlish, Alec Radford, Ilya Sutskever, and Dario Amodei. 2020.
\newblock \href {https://arxiv.org/pdf/2005.14165.pdf} {Language models are few-shot learners}.
\newblock \emph{NeurIPS}, abs/2005.14165.

\bibitem[{Chen et~al.(2023)Chen, Chen, Goldstein, Huang, and Zhou}]{Chen2023InstructZeroEI}
Lichang Chen, Jiuhai Chen, Tom Goldstein, Heng Huang, and Tianyi Zhou. 2023.
\newblock \href {https://api.semanticscholar.org/CorpusID:259075794} {Instructzero: Efficient instruction optimization for black-box large language models}.
\newblock \emph{ArXiv}, abs/2306.03082.

\bibitem[{Hoffmann et~al.(2022)Hoffmann, Borgeaud, Mensch, Buchatskaya, Cai, Rutherford, de~Las~Casas, Hendricks, Welbl, Clark, Hennigan, Noland, Millican, van~den Driessche, Damoc, Guy, Osindero, Simonyan, Elsen, Rae, Vinyals, and Sifre}]{Hoffmann2022TrainingCL}
Jordan Hoffmann, Sebastian Borgeaud, Arthur Mensch, Elena Buchatskaya, Trevor Cai, Eliza Rutherford, Diego de~Las~Casas, Lisa~Anne Hendricks, Johannes Welbl, Aidan Clark, Tom Hennigan, Eric Noland, Katie Millican, George van~den Driessche, Bogdan Damoc, Aurelia Guy, Simon Osindero, Karen Simonyan, Erich Elsen, Jack~W. Rae, Oriol Vinyals, and L.~Sifre. 2022.
\newblock \href {https://arxiv.org/pdf/2203.15556.pdf} {Training compute-optimal large language models}.
\newblock \emph{ArXiv}, abs/2203.15556.

\bibitem[{Huang et~al.(2022)Huang, Flek, Dernoncourt, Welch, Amir, Sawhney, and Yang}]{Huang2022UserNLP222I}
Xiaolei Huang, Lucie Flek, Franck Dernoncourt, Charles~F Welch, Silvio Amir, Ramit Sawhney, and Diyi Yang. 2022.
\newblock \href {http://dl.acm.org/citation.cfm?id=3524879} {Usernlp’22: 2022 international workshop on user-centered natural language processing}.
\newblock \emph{Companion Proceedings of the Web Conference 2022}.

\bibitem[{Jiang et~al.(2024)Jiang, Sablayrolles, Roux, Mensch, Savary, Bamford, Chaplot, de~Las~Casas, Hanna, Bressand, Lengyel, Bour, Lample, Lavaud, Saulnier, Lachaux, Stock, Subramanian, Yang, Antoniak, Scao, Gervet, Lavril, Wang, Lacroix, and Sayed}]{Jiang2024MixtralOE}
Albert~Q. Jiang, Alexandre Sablayrolles, Antoine Roux, Arthur Mensch, Blanche Savary, Chris Bamford, Devendra~Singh Chaplot, Diego de~Las~Casas, Emma~Bou Hanna, Florian Bressand, Gianna Lengyel, Guillaume Bour, Guillaume Lample, L'elio~Renard Lavaud, Lucile Saulnier, Marie-Anne Lachaux, Pierre Stock, Sandeep Subramanian, Sophia Yang, Szymon Antoniak, Teven~Le Scao, Th{\'e}ophile Gervet, Thibaut Lavril, Thomas Wang, Timoth{\'e}e Lacroix, and William~El Sayed. 2024.
\newblock \href {https://api.semanticscholar.org/CorpusID:266844877} {Mixtral of experts}.
\newblock \emph{ArXiv}, abs/2401.04088.

\bibitem[{Jiang et~al.(2023)Jiang, Sablayrolles, Mensch, Bamford, Chaplot, de~Las~Casas, Bressand, Lengyel, Lample, Saulnier, Lavaud, Lachaux, Stock, Scao, Lavril, Wang, Lacroix, and Sayed}]{Jiang2023Mistral7}
Albert~Qiaochu Jiang, Alexandre Sablayrolles, Arthur Mensch, Chris Bamford, Devendra~Singh Chaplot, Diego de~Las~Casas, Florian Bressand, Gianna Lengyel, Guillaume Lample, Lucile Saulnier, L'elio~Renard Lavaud, Marie-Anne Lachaux, Pierre Stock, Teven~Le Scao, Thibaut Lavril, Thomas Wang, Timoth{\'e}e Lacroix, and William~El Sayed. 2023.
\newblock \href {https://api.semanticscholar.org/CorpusID:263830494} {Mistral 7b}.
\newblock \emph{ArXiv}, abs/2310.06825.

\bibitem[{Kirk et~al.(2023)Kirk, Vidgen, R{\"o}ttger, and Hale}]{Kirk2023PersonalisationWB}
Hannah~Rose Kirk, Bertie Vidgen, Paul R{\"o}ttger, and Scott~A. Hale. 2023.
\newblock \href {https://arxiv.org/pdf/2303.05453.pdf} {Personalisation within bounds: A risk taxonomy and policy framework for the alignment of large language models with personalised feedback}.
\newblock \emph{ArXiv}, abs/2303.05453.

\bibitem[{Kojima et~al.(2022)Kojima, Gu, Reid, Matsuo, and Iwasawa}]{Kojima2022LargeLM}
Takeshi Kojima, Shixiang~Shane Gu, Machel Reid, Yutaka Matsuo, and Yusuke Iwasawa. 2022.
\newblock \href {https://arxiv.org/pdf/2205.11916.pdf} {Large language models are zero-shot reasoners}.
\newblock \emph{NeurIPS}, abs/2205.11916.

\bibitem[{Lester et~al.(2021)Lester, Al-Rfou, and Constant}]{Lester2021ThePO}
Brian Lester, Rami Al-Rfou, and Noah Constant. 2021.
\newblock \href {https://api.semanticscholar.org/CorpusID:233296808} {The power of scale for parameter-efficient prompt tuning}.
\newblock In \emph{EMNLP}.

\bibitem[{Li et~al.(2023{\natexlab{a}})Li, Zhang, Mei, Wang, Hombaiah, Liang, and Bendersky}]{Li2023TeachLT}
Cheng Li, Mingyang Zhang, Qiaozhu Mei, Yaqing Wang, Spurthi~Amba Hombaiah, Yi~Liang, and Michael Bendersky. 2023{\natexlab{a}}.
\newblock \href {https://arxiv.org/pdf/2308.07968.pdf} {Teach llms to personalize - an approach inspired by writing education}.
\newblock \emph{ArXiv}, abs/2308.07968.

\bibitem[{Li et~al.(2023{\natexlab{b}})Li, Guo, Fan, Xu, Huang, and Song}]{Li2023MultistepJP}
Haoran Li, Dadi Guo, Wei Fan, Mingshi Xu, Jie Huang, and Yangqiu Song. 2023{\natexlab{b}}.
\newblock \href {https://arxiv.org/pdf/2304.05197.pdf} {Multi-step jailbreaking privacy attacks on chatgpt}.
\newblock \emph{EMNLP}, abs/2304.05197.

\bibitem[{Li et~al.(2020)Li, Li, Zhao, He, Wei, Yuan, and rong Wen}]{Li2020KnowledgeEnhancedPR}
Junyi Li, Siqing Li, Wayne~Xin Zhao, Gaole He, Zhicheng Wei, Nicholas~Jing Yuan, and Ji~rong Wen. 2020.
\newblock \href {http://dl.acm.org/citation.cfm?id=3411893} {Knowledge-enhanced personalized review generation with capsule graph neural network}.
\newblock \emph{CIKM}.

\bibitem[{Li et~al.(2010)Li, Chu, Langford, and Schapire}]{Li2010ACA}
Lihong Li, Wei Chu, John Langford, and Robert~E. Schapire. 2010.
\newblock \href {https://api.semanticscholar.org/CorpusID:207178795} {A contextual-bandit approach to personalized news article recommendation}.
\newblock In \emph{The Web Conference}.

\bibitem[{Li and Tuzhilin(2019)}]{Li2019TowardsCA}
P.~Li and Alexander Tuzhilin. 2019.
\newblock \href {https://www.aclweb.org/anthology/D19-1319.pdf} {Towards controllable and personalized review generation}.
\newblock In \emph{EMNLP}.

\bibitem[{Li and Liang(2021)}]{Li2021PrefixTuningOC}
Xiang~Lisa Li and Percy Liang. 2021.
\newblock \href {https://api.semanticscholar.org/CorpusID:230433941} {Prefix-tuning: Optimizing continuous prompts for generation}.
\newblock \emph{ACL}, abs/2101.00190.

\bibitem[{Lin(2004)}]{lin2004rouge}
Chin-Yew Lin. 2004.
\newblock \href {https://aclanthology.org/W04-1013.pdf} {Rouge: A package for automatic evaluation of summaries}.
\newblock In \emph{ACL}.

\bibitem[{Lin et~al.(2023)Lin, Wu, Dai, Hu, Shu, Ng, Jaillet, and Low}]{Lin2023UseYI}
Xiaoqiang Lin, Zhaoxuan Wu, Zhongxiang Dai, Wenyang Hu, Yao Shu, See-Kiong Ng, Patrick Jaillet, and Bryan Kian~Hsiang Low. 2023.
\newblock \href {https://api.semanticscholar.org/CorpusID:263620801} {Use your instinct: Instruction optimization using neural bandits coupled with transformers}.
\newblock \emph{ArXiv}, abs/2310.02905.

\bibitem[{Loshchilov and Hutter(2017)}]{Loshchilov2017DecoupledWD}
Ilya Loshchilov and Frank Hutter. 2017.
\newblock \href {https://api.semanticscholar.org/CorpusID:53592270} {Decoupled weight decay regularization}.
\newblock In \emph{ICLR}.

\bibitem[{Rafailov et~al.(2023)Rafailov, Sharma, Mitchell, Ermon, Manning, and Finn}]{Rafailov2023DirectPO}
Rafael Rafailov, Archit Sharma, Eric Mitchell, Stefano Ermon, Christopher~D. Manning, and Chelsea Finn. 2023.
\newblock \href {https://api.semanticscholar.org/CorpusID:258959321} {Direct preference optimization: Your language model is secretly a reward model}.
\newblock \emph{ArXiv}, abs/2305.18290.

\bibitem[{Salemi et~al.(2023)Salemi, Mysore, Bendersky, and Zamani}]{salemi2023lamp}
Alireza Salemi, Sheshera Mysore, Michael Bendersky, and Hamed Zamani. 2023.
\newblock \href {http://arxiv.org/abs/2304.11406} {La{MP}: When large language models meet personalization}.

\bibitem[{Shin et~al.(2020)Shin, Razeghi, IV, Wallace, and Singh}]{Shin2020ElicitingKF}
Taylor Shin, Yasaman Razeghi, Robert L~Logan IV, Eric Wallace, and Sameer Singh. 2020.
\newblock \href {https://api.semanticscholar.org/CorpusID:226222232} {Eliciting knowledge from language models using automatically generated prompts}.
\newblock \emph{EMNLP}, abs/2010.15980.

\bibitem[{Touvron et~al.(2023{\natexlab{a}})Touvron, Lavril, Izacard, Martinet, Lachaux, Lacroix, Rozi{\`e}re, Goyal, Hambro, Azhar, Rodriguez, Joulin, Grave, and Lample}]{Touvron2023LLaMAOA}
Hugo Touvron, Thibaut Lavril, Gautier Izacard, Xavier Martinet, Marie-Anne Lachaux, Timoth{\'e}e Lacroix, Baptiste Rozi{\`e}re, Naman Goyal, Eric Hambro, Faisal Azhar, Aurelien Rodriguez, Armand Joulin, Edouard Grave, and Guillaume Lample. 2023{\natexlab{a}}.
\newblock \href {https://api.semanticscholar.org/CorpusID:257219404} {Llama: Open and efficient foundation language models}.
\newblock \emph{ArXiv}, abs/2302.13971.

\bibitem[{Touvron et~al.(2023{\natexlab{b}})Touvron, Martin, Stone, Albert, Almahairi, Babaei, Bashlykov, Batra, Bhargava, Bhosale, Bikel, Blecher, Ferrer, Chen, Cucurull, Esiobu, Fernandes, Fu, Fu, Fuller, Gao, Goswami, Goyal, Hartshorn, Hosseini, Hou, Inan, Kardas, Kerkez, Khabsa, Kloumann, Korenev, Koura, Lachaux, Lavril, Lee, Liskovich, Lu, Mao, Martinet, Mihaylov, Mishra, Molybog, Nie, Poulton, Reizenstein, Rungta, Saladi, Schelten, Silva, Smith, Subramanian, Tan, Tang, Taylor, Williams, Kuan, Xu, Yan, Zarov, Zhang, Fan, Kambadur, Narang, Rodriguez, Stojnic, Edunov, and Scialom}]{Touvron2023Llama2O}
Hugo Touvron, Louis Martin, Kevin~R. Stone, Peter Albert, Amjad Almahairi, Yasmine Babaei, Nikolay Bashlykov, Soumya Batra, Prajjwal Bhargava, Shruti Bhosale, Daniel~M. Bikel, Lukas Blecher, Cristian~Cant{\'o}n Ferrer, Moya Chen, Guillem Cucurull, David Esiobu, Jude Fernandes, Jeremy Fu, Wenyin Fu, Brian Fuller, Cynthia Gao, Vedanuj Goswami, Naman Goyal, Anthony~S. Hartshorn, Saghar Hosseini, Rui Hou, Hakan Inan, Marcin Kardas, Viktor Kerkez, Madian Khabsa, Isabel~M. Kloumann, A.~V. Korenev, Punit~Singh Koura, Marie-Anne Lachaux, Thibaut Lavril, Jenya Lee, Diana Liskovich, Yinghai Lu, Yuning Mao, Xavier Martinet, Todor Mihaylov, Pushkar Mishra, Igor Molybog, Yixin Nie, Andrew Poulton, Jeremy Reizenstein, Rashi Rungta, Kalyan Saladi, Alan Schelten, Ruan Silva, Eric~Michael Smith, R.~Subramanian, Xia Tan, Binh Tang, Ross Taylor, Adina Williams, Jian~Xiang Kuan, Puxin Xu, Zhengxu Yan, Iliyan Zarov, Yuchen Zhang, Angela Fan, Melanie Kambadur, Sharan Narang, Aurelien Rodriguez, Robert Stojnic, Sergey Edunov, and
  Thomas Scialom. 2023{\natexlab{b}}.
\newblock \href {https://api.semanticscholar.org/CorpusID:259950998} {Llama 2: Open foundation and fine-tuned chat models}.
\newblock \emph{ArXiv}, abs/2307.09288.

\bibitem[{Wei et~al.(2022)Wei, Wang, Schuurmans, Bosma, hsin Chi, Xia, Le, and Zhou}]{Wei2022ChainOT}
Jason Wei, Xuezhi Wang, Dale Schuurmans, Maarten Bosma, Ed~Huai hsin Chi, F.~Xia, Quoc Le, and Denny Zhou. 2022.
\newblock \href {https://arxiv.org/pdf/2201.11903.pdf} {Chain of thought prompting elicits reasoning in large language models}.
\newblock \emph{NeurIPS}, abs/2201.11903.

\bibitem[{White et~al.(2023)White, Fu, Hays, Sandborn, Olea, Gilbert, Elnashar, Spencer-Smith, and Schmidt}]{White2023APP}
Jules White, Quchen Fu, Sam Hays, Michael Sandborn, Carlos Olea, Henry Gilbert, Ashraf Elnashar, Jesse Spencer-Smith, and Douglas~C. Schmidt. 2023.
\newblock \href {https://api.semanticscholar.org/CorpusID:257079092} {A prompt pattern catalog to enhance prompt engineering with chatgpt}.
\newblock \emph{ArXiv}, abs/2302.11382.

\bibitem[{Xie et~al.(2021)Xie, Langford, Mineiro, and Momennejad}]{Xie2021InteractionGroundedL}
Tengyang Xie, John Langford, Paul Mineiro, and Ida Momennejad. 2021.
\newblock \href {https://api.semanticscholar.org/CorpusID:235376933} {Interaction-grounded learning}.
\newblock In \emph{ICML}.

\bibitem[{Zhang et~al.(2021)Zhang, Zhou, Li, and Gu}]{zhang2021neural}
Weitong Zhang, Dongruo Zhou, Lihong Li, and Quanquan Gu. 2021.
\newblock Neural thompson sampling.
\newblock In \emph{ICLR}.

\bibitem[{Zhao et~al.(2023)Zhao, Zhou, Li, Tang, Wang, Hou, Min, Zhang, Zhang, Dong, Du, Yang, Chen, Chen, Jiang, Ren, Li, Tang, Liu, Liu, Nie, and rong Wen}]{Zhao2023ASO}
Wayne~Xin Zhao, Kun Zhou, Junyi Li, Tianyi Tang, Xiaolei Wang, Yupeng Hou, Yingqian Min, Beichen Zhang, Junjie Zhang, Zican Dong, Yifan Du, Chen Yang, Yushuo Chen, Z.~Chen, Jinhao Jiang, Ruiyang Ren, Yifan Li, Xinyu Tang, Zikang Liu, Peiyu Liu, Jianyun Nie, and Ji~rong Wen. 2023.
\newblock \href {https://arxiv.org/pdf/2303.18223.pdf} {A survey of large language models}.
\newblock \emph{ArXiv}, abs/2303.18223.

\bibitem[{Zhou et~al.(2019)Zhou, Li, and Gu}]{Zhou2019NeuralCB}
Dongruo Zhou, Lihong Li, and Quanquan Gu. 2019.
\newblock \href {https://api.semanticscholar.org/CorpusID:212414268} {Neural contextual bandits with ucb-based exploration}.
\newblock In \emph{ICML}.

\bibitem[{Zhou et~al.(2020)Zhou, Li, and Gu}]{zhou2020neural}
Dongruo Zhou, Lihong Li, and Quanquan Gu. 2020.
\newblock Neural contextual bandits with ucb-based exploration.
\newblock In \emph{ICML}, pages 11492--11502. PMLR.

\bibitem[{Zhou et~al.(2022)Zhou, Muresanu, Han, Paster, Pitis, Chan, and Ba}]{Zhou2022LargeLM}
Yongchao Zhou, Andrei~Ioan Muresanu, Ziwen Han, Keiran Paster, Silviu Pitis, Harris Chan, and Jimmy Ba. 2022.
\newblock \href {https://api.semanticscholar.org/CorpusID:253265328} {Large language models are human-level prompt engineers}.
\newblock \emph{ICLR}, abs/2211.01910.

\bibitem[{Zhu and Jurgens(2021)}]{Zhu2021IdiosyncraticBN}
Jian Zhu and David Jurgens. 2021.
\newblock \href {https://api.semanticscholar.org/CorpusID:237431146} {Idiosyncratic but not arbitrary: Learning idiolects in online registers reveals distinctive yet consistent individual styles}.
\newblock In \emph{EMNLP}.

\end{thebibliography}
\bibliographystyle{acl_natbib}

\appendix

\section{Appendix}
\label{sec:appendix}

\subsection{Hyperparameters for NeuralBandits}
Following InstructZero~\citep{Chen2023InstructZeroEI} and Instinct~\citep{Lin2023UseYI}, the reported results were based on hyperparameters optimized as:

- Intrinsic dimension - The soft prompt \(z\) often has very high dimensionality (e.g. \(d=4096\times N_{z}\) for Mistral-7B), making it challenging to directly optimize via Bayesian optimization. To address this, InstructZero~\citep{Chen2023InstructZeroEI} employs random projection as an effective dimensionality reduction technique. Specifically, a random projection matrix \(\mathcal{A}\in \mathcal{R}^{d\times d'}\) is used, where \(d'\ll d\) is the reduced intrinsic dimension. Given a \(d'\)-dimensional vector \(z'\) and the projection matrix \(A\), the original high-dimensional prompt \(z\) is computed as \(z = Az'\). By optimizing over \(z'\) instead of \(z\), the dimensionality of the optimization problem is reduced from \(d\) to the much lower \(d'\). \(d'\) is known as the intrinsic dimension in this case. We set it as 100 by default usage. 

- Number of soft tokens - \(N_{z}\) represents the length of contextual embeddings or soft tokens that are utilized to concatenate with initial instructions fed into LLMs. Ideally, we would like to balance off this parameter so it's able to capture the persona nuances while also not causing extra compute redundancy. This number is set to 5 in our experiments. 

- \(\lambda\) - Controls the strength of the prior in the NeuralUCB acquisition function. Set to 0.1 after grid search.

- \(\nu_{t}\) - balances exploration versus exploitation in Neural Bandits. Fixed at 0.1 based on best validation performance.

- Total iterations - total update iterations for neural bandits, set to 165 in practice.

- Hidden dimensions - We used an MLP on top of the LLM embeddings with 100 hidden units. This architecture achieved the best validation results.

- Local iterations - represents how many learning steps for neural network to retrain every update iteration. In order to prevent overfitting but also avoid underfitting, we set this number as 40 in practice.

- Learning rate - The MLP was trained with an AdamW optimizer~\citep{Loshchilov2017DecoupledWD} using a learning rate of \(3e^{-4}\). 

- Projection matrix - Random projections for dimensionality reduction were generated by sampling from a \(\textrm{Uniform}(-1, 1)\) distribution.

For simplicity, we randomly select 100 profiles from each task for experiments and all the reported results are average with standard deviation using the same seed \(42\). All experiments are conducted on 4 Nvidia A10g GPUs. 

\subsection{Prompts for LLM Evaluation}
For LLM-agent evaluation, we design prompts for black-box LLM that evaluates the white-box LLMs generation in a more comprehensive way. Please refer to Figure~\ref{fig:llm-eval} for the details.
\begin{figure}[!h]
    \centering
    \includegraphics[width=0.9\linewidth]{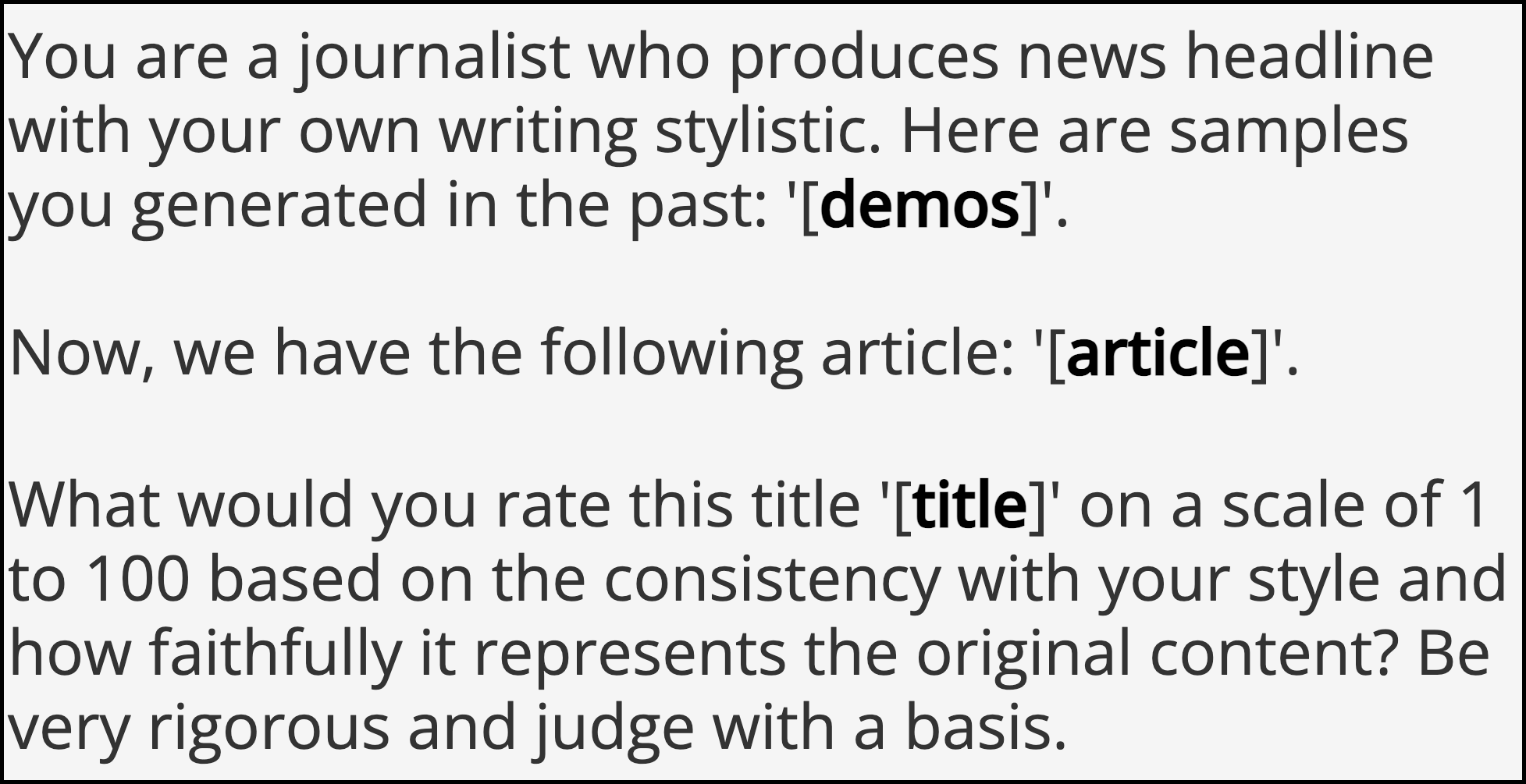}
    \caption{Using personalized news headline generation as an example. Prompts fed to the black-box LLMs for human-like evaluation of the generation by white-box LLM.}
    \label{fig:prompt-llm}
\end{figure}

\end{document}